\pdfoutput=1

\documentclass[11pt]{article}

\usepackage[]{acl}

\usepackage{times}
\usepackage{latexsym}

\usepackage{amsfonts}
\usepackage{graphicx}
\usepackage{bigstrut}
\usepackage{multirow}
\usepackage{booktabs}
\usepackage{amsthm}
\usepackage{mathrsfs}
\usepackage{arydshln}
\usepackage{algorithm}
\usepackage{algorithmic}
\usepackage[T1]{fontenc}

\usepackage[utf8]{inputenc}

\usepackage{microtype}
\usepackage{bm}

\usepackage{amsmath,amsthm,amsfonts,amssymb}
\usepackage{bm}
\usepackage{graphicx}
\usepackage{sidecap}
\usepackage{mathrsfs}
\usepackage{multirow, multicol}
\usepackage{caption}

\usepackage{wrapfig}
\usepackage{booktabs}

\usepackage{algorithm}
\usepackage{algorithmic}
\usepackage{natbib} 

\newcommand{\bbR}{\mathbb{R}}

\newcommand{\calA}{\mathcal{A}}

\newcommand{\calD}{\mathcal{D}}
\newcommand{\calE}{\mathcal{E}}

\newcommand{\calG}{\mathcal{G}}

\newcommand{\calL}{\mathcal{L}}

\newcommand{\calP}{\mathcal{P}}
\newcommand{\calQ}{\mathcal{Q}}
\newcommand{\calR}{\mathcal{R}}

\def\1{\bm{1}}

\def\vb{{\bm{b}}}
\def\vc{{\bm{c}}}

\def\ve{{\bm{e}}}

\def\vp{{\bm{p}}}
\def\vq{{\bm{q}}}

\def\vx{{\bm{x}}}

\def\mA{{\bm{A}}}

\def\mC{{\bm{C}}}

\def\mX{{\bm{X}}}

\DeclareMathAlphabet{\mathsfit}{\encodingdefault}{\sfdefault}{m}{sl}
\SetMathAlphabet{\mathsfit}{bold}{\encodingdefault}{\sfdefault}{bx}{n}

\title{Semi-constraint Optimal Transport \\for Entity Alignment with Dangling Cases}

\author{
Shengxuan Luo$^{1}$ \space\space
Pengyu Cheng$^{2}$ \space\space
Sheng Yu$^{1}$ \space\space\\
$^{1}$Tsinghua University \space\space\space\space
$^{2}$Duke University \\
\texttt{luosx18@mails.tsinghua.edu.cn}
}

\usepackage{amsmath}

\begin{document}
\maketitle
\begin{abstract}
\vspace{-2.mm} 

Entity alignment (EA)  merges  knowledge graphs (KGs) by identifying the equivalent entities in different graphs, which can effectively enrich knowledge representations of KGs.
%
However, in practice, different KGs often include \textit{dangling entities} whose counterparts cannot be found in the other graph, which limits the performance of EA methods. 
%
%
To improve EA with dangling entities, we propose an unsupervised method called \textbf{S}emi-constraint \textbf{O}ptimal \textbf{T}ransport for \textbf{E}ntity \textbf{A}lignment in \textbf{D}angling cases (\textbf{SoTead}).
Our main idea is to model the entity alignment between two KGs as an optimal transport  problem from one KG's entities to the others. First, we set pseudo entity pairs between KGs based on pretrained word embeddings. Then, we conduct contrastive metric learning to obtain the transport cost between each entity pair. Finally, we introduce a \textit{virtual entity} for each KG to ``align'' the dangling entities from the other KGs, which relaxes the optimization constraints and leads to a semi-constraint optimal transport.
In the experimental part, we first show the superiority of SoTead on a commonly-used entity alignment dataset. Besides, to analyze the ability for dangling entity detection with other baselines, we construct a medical cross-lingual knowledge graph dataset, MedED, where our SoTead also reaches state-of-the-art performance.
\end{abstract}


\vspace{-3mm} 
\section{Introduction}
\vspace{-1.5mm} 
Knowledge graphs (KGs) has become one of the essential modules in various intelligent  domains, such as question-answering~\citep{savenkov2016crqa,yu2017improved,jin2021biomedical}, recommendation~\citep{cao2019unifying}, and search engines~\citep{xiong2017explicit}. However, constructing a large-scale KG consumes massive computational resources \citep{paulheim2018much} and can be easily trapped into knowledge incompleteness \citep{galarraga2017predicting}. Therefore, under real-world scenarios, merging multiple KGs by entity alignment (EA) becomes a mainstream solution to enrich knowledge of KGs. More specifically, entity alignment combines different KGs  by aligning  the equivalent entities across the KGs \citep{zhang2020industry,zeng2021comprehensive}. 

To detect the identical entities, prior works of entity alignment mainly focus on embedding-level alignment by learning a representation vector for each entity, so that the similar entities can locate closely to each others in the embedding spaces~\citep{chen2017multilingual,sun2018bootstrapping,wang2018cross,zhu2021raga,liu2021self,lin2021echoea}. These embedding-based EA methods always require an  impractical assumption, that for each entity there always exists a counterpart in the other KG~\citep{sun2021knowing}. Moreover, the embedding-level EA  methods evaluate their performances only on the existing entity pairs between the testing KGs. However, in various real-world scenarios, the identical pair of one entity is not guaranteed to exist between different KGs \citep{zhao2020towards,sun2021knowing}, which limits the application range of embedding-based EA methods.

Recently, the aforementioned unmatchable entities in KGs have been recognized as the \textit{dangling entities}~\citep{sun2021knowing}. To improve the EA perfomance, it is necessary to identify the dangling entities and then align the remaining matchable entities to their counterparts. Hence, a limited number of prior works attempt to address the dangling entities detection (DED). \citet{zhao2020experimental} and \citet{zeng2021towards} apply thresholds on the embedding distance of each entity and its nearest neighbor to determine if the entity is dangling. Nevertheless, the threshold cuts are not flexible enough when considering the relative entity distances. For example, a dangling entity could have an available distance (smaller than the threshold) to a target entity, while the target one has even closer distance to the ground-truth counterpart.
%

%
%
%
\citet{sun2021knowing} constructs a labeled training set of dangling entities and trains a neural classifier to predict whether each entity is matchable. However, this supervised method has limited practicability for the  difficulty of creating training sets in real-world large-scale KGs.

 To solve the dangling entities within an unsupervised scheme and a global perspective,
we propose \textbf{SoTead}, a \textbf{S}emi-constraint \textbf{O}ptimal \textbf{T}ransport method for \textbf{E}ntity \textbf{A}lignment with \textbf{D}angling cases. ((Instead of matching the entities with their closest neighbors)), we consider a global optimal entity matching by solving the optimal transport between the groups of entities in two KGs. Our unsupervised method consists of two parts: (1) estimating the cross-KG entity similarity with the textual information of  entity names  and a contrastive distance learning in entity embedding space; (2) solving a \textit{semi-constraint} optimal transport (OT) between two KG entity sets to discover the globally optimal entity matching based on the estimated entity similarities. Since the dangling entities cannot ``transport'' to the target KG, we introduce a virtual entity in the target KG and relax the OT constraints, so that the dangling entities can be matched to the virtual entity instead. On the other hand, the virtual entity also works as a dangling entities detector in the solution of our semi-constraint OT optimization, for only dangling entities in the target KG with match to it. 
To empirically demonstrate the effectiveness of our SoTead method, we conduct the experiments on both the entity alignment (EA)  and dangling entity detection (DED) tasks. Since there is no testing set for dangling entity detection, we also construct a cross-lingual medical knowledge graph dataset, MedED.
Our experiments show that the dangling entity identification mechanism also enhances the EA performance. On both of the EA and DED tasks, our method achieves  remarkable performance improvement even comparing with baselines under supervised setups.
%
%
%
%
%

The main contributions of this paper is summarized in follows:
\begin{itemize}
\vspace{-1.mm} 
    \item We propose SoTead, an unsupervised entity alignment method via semi-constraint optimal transport, which can jointly match paired entities and detect unmatchable entities. 
   \vspace{-1.mm} 
	\item  To demonstrate the effectiveness of our model, we create a cross-lingual knowledge graph dataset, MedED, which supports the evaluation for both entity alignment and dangling entity detection.
	\vspace{-1.mm} 
	\item We achieve state-of-the-art performance on  comprehensive experiments of entity alignment and dangling entity detection, and analyze the impacts of high-quality dangling entity detection for improving entity alignment.
\end{itemize}

The source code is publicly available at
 \url{https://github.com/luosx18/UED}.

\vspace{-3.mm} 
\section{Preliminary}\label{sec:pre-ot}\label{sec:MIP}
\vspace{-1.5mm} 
\textbf{Optimal Transport: }
Optimal Transport (OT) aims to find the minimal cost for transporting the density distribution of  group of items to that of another group, given all the unit transportation cost between each item pairs \citep{villani2009optimal}. In formula, OT between density $\vp \in \bbR^n$ and $\vq \in \bbR^m$ is:
\begin{equation}\begin{aligned}
D_{\mC} (\vp, \vq):=& \min_{\mX}  \langle \mC, \mX\rangle, \\
\text{s.t.}  &\mX^T \bm{1}_m = \vp,  \mX \bm{1}_n = \vq, \\
& \vq^T \bm{1}_n = \vp^T \bm{1}_m,
\end{aligned}\end{equation} 
where $\langle\mC, \mX\rangle = \sum_{i=1}^{n}\sum_{j=1}^{m} c_{ij}x_{ij}$ is the matrix inner product between cost matrix $\mC= [c_{ij}]_{m\times n} $ and the transited density $\mX = [x_{ij}]_{m\times n} $, and $\bm{1}_m$, $\bm{1}_n$ are all-one vectors.

Since the OT objective is symmetric for both density $\vq$ and $\vp$,  OT has  been recognized as an effective match approach and shown noticeable matching performance in many machine leaning tasks. For example, in machine translation,  \citet{alvarez2018gromov} uses OT to measure the similarity of generated translation with groud-truth, by matching sentences as two groups of words. Besides, OT has shown improvements on other natural language processing tasks, such as  document retrieval \citep{kusner2015word}, text generation \citep{salimans2018improving,chen2018adversarial}, and sequence-to-sequence learning \citep{chen2019improving}.

\noindent \textbf{Mixed Integer Programming: }
Linear programming (LP) is a fundamental method to optimize  linear objective functions with linear constraints \citep{padberg1991branch,gass2003linear,dantzig2016linear}. Mathematically, LP is formed as:
\begin{equation}\begin{aligned}
\min \ \  &  \mathbf{c}^{T} \mathbf{x}, \\
\text{s.t.  } & \mathbf{A} \mathbf{x} \leq \mathbf{b},\\
& \mathbf{x}\geq \mathbf{0},
\end{aligned}\end{equation}
where $\bm{x} = (x_1, x_2, \dots, x_n)^T$ is the vector of $n$ variables, $\mathbf{A} \mathbf{x} \leq \mathbf{b}$ ($\mA\in \bbR^{m \times n}, \vb\in \bbR^{m}$) are $m$ linear constrains, and $\vc^T \vx$ is the linear objective.
When considering the linear programming with continuous variables, the simplex algorithm is widely-used, which efficiently searches the optimal solution on the vertices of the feasible region \citep{dantzig1955generalized}. 
However, when limiting some of the variables to be integers, the linear programming becomes more challenging, leading to the problem of mixed integer programming (MIP).
%

 A common solution to MIP is the branch-and-cut algorithm \citep{wolsey1999integer,mitchell2002branch}, which first solves the relaxed linear programming without the integer constraints via the simplex method, and then iteratively branches the MIP by splitting the constraint space based on one of the non-integer variables. For every branch, a relaxed linear programming is solved to check if the relaxation is infeasible. If the relaxation is infeasible or less optimal than current solution, the corresponding branch will be pruned. Otherwise, the branch process will continue until all the constraints are satisfied (details in the Supplementary~\ref{sec:branch-and-cut}).
 %
 %
 Besides, \citet{balas1993lift,mitchell2002branch} accelerate the solving of the MIP via  a cutting plane algorithm for constraint branching.

\vspace{-1.5mm} 
\section{Proposed Method}
\vspace{-1.5mm} 
\label{sec:section 3}
\label{sec:proposed-method}
In this section, we first jointly model the entity alignment and  dangling entity detection within a semi-constraint optimal transport framework. Then we describe how to obtain the entity distances across the KGs under an unsupervised setup. At last, we analyze the numerical solution and efficiency of the semi-constraint optimal transport.


\vspace{-1.5mm}
\subsection{Semi-constraint Optimal Transport}
\vspace{-1.mm}
We define a knowledge graph as  $\mathcal{G}=\left\{\mathcal{E},\mathcal{R}\right\}$, where $\calE$ is the entity set, and  $\mathcal{R}$ denotes the set of entity relations. Furthermore, we consider the entity set $\mathcal{E}=\mathcal{D}\cup\mathcal{A}$ as a disjoint union of the dangling set $\mathcal{D}$ and matchable set $\mathcal{A}$.
For two KGs, $\mathcal{G}_1 = \{\calD_1 \cup \calA_1, \calR_1 \}$ and $\mathcal{G}_2 = \{\calD_2 \cup \calA_2, \calR_2 \}$, our task is to align the matchable entities in  $\mathcal{A}_1$ and $\mathcal{A}_2$ and detect the dangling set  $\mathcal{D}_1$ and $\mathcal{D}_2$ jointly.
%

Denote entities in $\calG_1$ as $\calE_1 = \{u_1, u_2, \dots, u_m\}$ and entities in $\calG_2$ as $\calE_2 = \{v_1, v_2, \dots, v_n\}$. Then entity alignment is to find equivalent pairs $\{ (u_i, v_j) | u_i \in \calA_1, v_j \in \calA_2\}$. As in Section~\ref{sec:pre-ot}, we convert this matching problem to an optimal transport from $\calE_1$ to $\calE_2$ by minimizing the dissimilarity scores $\{d(u_i, v_j)\}$ as the transportation cost. Hence, we introduce indicators:
\begin{equation}
    \begin{aligned}
    \psi_{ij} = \begin{cases}
         1,& \text{if $u_i$ matches $v_j$} \\
         0,& \text{if $u_i$ does not  match $v_j$} 
    \end{cases}
    \end{aligned}
\end{equation}
With the indicators $\psi_{ij}$, we convert the global entity alignment in to the following objective:
\begin{equation}\label{eq:naive-motivation}
\begin{aligned}
\min &\sum_{i=1}^m \sum_{j=1}^{n} c_{ij}\psi_{ij},\\
\text{s.t.} &\sum\limits_{j=1}^n\psi_{ij}=1, \ \ 1\leq i\leq m,\\
&\sum\limits_{i=1}^m\psi_{ij}=1, \  \ 1\leq j\leq n,
\end{aligned}
\end{equation}
where the cost $c_{ij} = d(u_i, v_j)>0$ is set to the dissimilarity between entity pair $(u_i , v_j)$. The summation constraints indicate each entity in $\calE_1$ can only match to one entity in $\calE_2$ and vice versa.
%

However, when  $\calE_1$ and $\calE_2$ have different numbers of entities ($n\neq m$), the constraints in equation~\eqref{eq:naive-motivation} can never be satisfied. Besides, the dangling entities are not considered in the above optimal transport.
%
%
To address these issues, we introduce ``virtual'' entities $u_0$ and $v_0$ into  $\mathcal{E}_1$ and $\mathcal{E}_2$ respectively. With the two virtual entities $u_0$ and $v_0$, we can further relax the constraints of indicators:
\begin{align}
\min &\sum_{i=1}^m \sum_{j=1}^n c_{ij}\psi_{ij} + \sum_{i=1}^m c_{i,0} \psi_{i,0} + \sum_{j=1}^n c_{0,j} \psi_{0,j} \nonumber\\
\text{s.t.} &\sum_{j=1}^n \psi_{ij}=1, \ \ 1\leq i\leq m, \label{eq:relaxed-ot}\\
&\sum_{i=1}^m \psi_{ij}=1, \ \ 1\leq j\leq n. \nonumber
\end{align} 
Since no marginal constraint is added to $\{\psi_{0,j}\}_{j=1}^n$ and $\{\psi_{i,0}\}_{i=1}^m$, arbitrary number of dangling entities can be aligned to $u_0$ and $v_0$, which means the constrains in equation~\eqref{eq:relaxed-ot} are always feasible. Note that we do not consider the distance $c_{0,0}= d(u_0, v_0)$ between the two virtual entities, because the alignment $\psi_{0,0}$ between $u_0$ and $v_0$ is meaningless and can lead to the degeneration of the optimization.

Also, the distances between  virtual entities and real entities have not been defined. Here, we introduce two hyper-parameters $\alpha,\beta > 0$, as the cross KG distances between the virtual entities and real entities, $c_{0,j} = \alpha, c_{i,0} = \beta$ ($1\leq j \leq n$ and $1 \leq i \leq m$). The details about the selection of $\alpha$ and $\beta$ are described in Section~\ref{sec:cost-function}. Therefore, the final optimization of our semi-constraint optimal transport is in equation~\eqref{eq:final-obj}.
By solving equation~\eqref{eq:final-obj}, we jointly obtain the solutions for entity alignment (with $\{\psi_{ij}\}_{1 \leq i \leq m, 1 \leq j \leq n}$ ) and dangling entity detection (with $\{\psi_{0,j}\}_{j=1}^n$ and $\{\psi_{i,0}\}_{i=1}^m$). In the followings, we first describe how we obtain the entity distances $c_{ij}$ in Section~\ref{sec:cost-function}, then discuss how to solve the semi-constraint optimal transport in Section~\ref{sec:solving-sot}.
\begin{align}
\min &\sum_{i=1}^m \sum_{j=1}^n c_{ij}\psi_{ij} + \beta \sum_{i=1}^m \psi_{i,0} + \alpha \sum_{j=1}^n \psi_{0,j} \nonumber\\
\text{s.t.} &\sum_{j=1}^n \psi_{ij}=1, \ \ 1\leq i\leq m, \label{eq:final-obj}\\
&\sum_{i=1}^m \psi_{ij}=1, \ \ 1\leq j\leq n. \nonumber 
\end{align} 

\vspace{-1.mm} 
\subsection{Entity Distance Learning} \label{sec:cost-function}
\begin{figure}[t]
    \centering
    \includegraphics[scale=0.415]{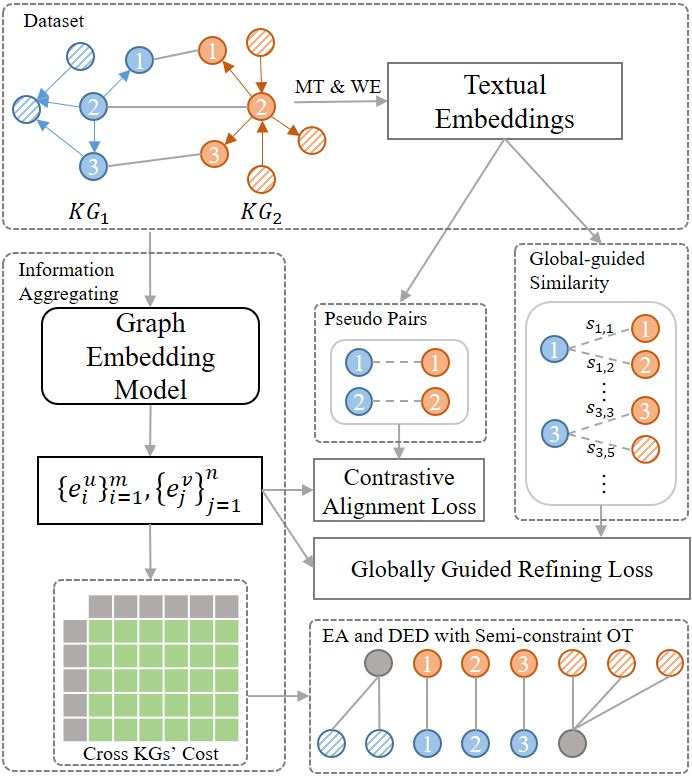}
    \vspace{-1mm}
    \caption{The Framework of SoTead. 
    The circles with a number are matchable entities, and the circles with slash denote dangling entities. The gray circles are the virtual entities and the gray rectangles in distance matrix denotes distance between virtual entity to other entities. MT and WE refer to machine translation and word embeddings.}
    \label{fig:sotead-framework}
\vspace{-1mm}
\end{figure}

As shown in Figure~\ref{fig:sotead-framework}, our unsupervised entity distance learning scheme consists of three steps: (1) set a group of pseudo entity pairs based on the textual information of entity names; (2) with the guidance of pseudo pairs, conduct a contrastive embedding learning on entity sample pairs; (3) utilize the textual embedding of all entities to build a globally guided refining loss. The details of the three steps is described below.


\textbf{Pseudo entity pairs:} We utilize GloVe \citep{pennington2014glove} word embeddings to obtain entity name textual embeddings  $\bm{w}^u_i, \bm{w}^v_j$ for entities $u_i \in \calE_1 $ and $v_i \in \calE_2$ respectively. Then the initial similarity is defined as $s_{ij} = \cos(\bm{w}^u_i, \bm{w}^v_j)$. 
By setting a threshold $0< \varepsilon<1$, we select pseudo entity pairs when  $s_{ij}$ satisfies: (a) $s_{ij}> \varepsilon$; (b) $s_{ik} \leq \varepsilon$, for any $k\neq j$; (c) $s_{lj} \leq \varepsilon$, for any $l \neq i$.
We denote the pseudo entity pairs set as $\mathcal{P}$. For cross-lingual KGs, we translate entity names via machine translation before extracting textual information.

\textbf{Contrastive alignment loss:}
We use the Pseudo entity pair to induce the learning of  enhanced entity embeddings $\bm{e}^{u}_i$ and $\bm{e}^{v}_j$ for $u_i$ and $v_j$ respectively. Each entity embedding is encoded with a graph neural network (GNN), which takes the textual embedding as node embedding and KG relations as the graph edges.
For each pseudo pair $(u_i, v_j) \in \calP$, we can define the negative pair set $\bar{\calP}(u_i, v_j)$ by replacing $e_i$ or $e_j$ with their neighbors \citep{mao2020relational,zhu2021relation}. Then
we use the following contrastive loss with negative sample pair $(u_{i'}, v_{j'}) \in \bar{\calP}(u_i, v_j)$:
\begin{equation*}
\begin{aligned}
 \sum_{i', j'}\max\{d_M(\ve^u_i,\ve^v_j)-d_M\left(\ve^u_{i'},\ve^v_{j'}\right)+\lambda), 0\}
\end{aligned}
\end{equation*}
where $\lambda$ is the margin, and $d_M(\cdot,\cdot)$ is the Manhattan distance following previous works \citep{zhu2021raga}. The overall alignment loss $\calL_a$ is the summation over all pseudo pairs and their negative sample pairs.

\textbf{Globally guided refining loss:}
The alignment loss $\calL_a$ only trains pseudo entity pairs and their negative samples. To make full use of entity textual information and learn embedding for all entities, we add a globally guided refining loss. We denote $\mathcal{Q}$ as  all $(u_i, v_j)$ pairs, where $v_j$ is one of the top $N$ similar entities of $u_i$ based on textual similarity $\{s_{ij}\}$, and $N$ is a hyper-parameter. Negative sample set $\bar{\calQ}(u_i, v_j)$ also replaces $e_i$ or $e_j$ with their neighbors. For each $(u_i, v_j) \in \calQ$ and its all negative samples $(u_{i'}, v_{j'}) \in \bar{\calQ}(u_i, v_j)$, we set:
\begin{equation*}
s_{ij} \sum_{i',j'}\max \{d_M(\ve^u_i,\ve^v_j)-d_M\left(\ve^u_{i'},\ve^v_{j'}\right)+\lambda,0\}\end{equation*}
Here we multiply $s_{ij}$ to the contrastive loss, to induce the model paying more attention to the entity pairs with higher textual similarity. The overall globally guided loss $\calL_g$ is the summation over all entity pairs and their negative sample pairs.

 
 The overall loss for entity embeddings is
\begin{equation}
\mathcal{L} = \mathcal{L}_a+w(t)\mathcal{L}_g,
\end{equation} 
where $t$ is the training step, and $w(t)$ decreases linearly to 0 as $t$ increases. When entity embeddings $\ve^u_i$ and $\ve^v_j$ are learned, the cost for our optimal transport is set as $c_{ij} = d_M (\ve^u_i, \ve^v_j)$. 

For the cost for transporting entities to virtual entities, 
 we first denote $l_{i}^u =\text{min}_{1\leq j\leq n} \{c_{ij}\}$ and $l_{j}^v =\text{min}_{1\leq i\leq m} \{c_{ij}\}$ as the smallest cost from $u_i$ to $\mathcal{G}_2$ and from $v_j$ to $\mathcal{G}_1$, respectively. Next, we do grid-search of the quantiles of $\{l_{i}^u, 1\leq i \leq m\}$ and $\{l_{j}^v, 1\leq j \leq n\}$ in pairs as possible selection of $\alpha$ and $\beta$. The optimal selection of $\alpha^*$ and $\beta^*$ achieves the best EA performance on the pseudo entity pairs set $\mathcal{P}$.

\vspace{-1.5mm}
\subsection{Integer Programming Solution}\label{sec:solving-sot}
\vspace{-.5mm}
%
To solve the semi-constraint optimal transport in equation~\eqref{eq:final-obj}, we treat it as a mixed integer programming(MIP) as described in Section~\ref{sec:MIP}. 
%
Then we use a branch-and-cut algorithm to calculate the indicator solutions. The details of solution steps for equation~\eqref{eq:final-obj} is described in Supplementary~\ref{sec:branch-and-cut}.

Moreover, to further reduce the complexity of the semi-constraint optimal transport, we only consider the transports from each entity to its top-$K$ similar entities in the target KG, noting the fact that less similar entities contain less information for alignment. This approximation also leads to more sparse sets of $c_{ij}$ and $\psi_{ij}$. In Section~\ref{sec:section 5.3}, we show that this simplification is powerful with less  computational complexity.


\vspace{-1mm}
\section{Related Work}
\label{sec:section 2}
\label{sec:related-work}
\vspace{-1mm}
\textbf{Embedding-based Entity Alignment: } The aim of 
embedding-based entity alignment is 
to encode KGs into low-dimensional vector spaces, then find nearest neighbors as the aligned entities. 
Recently various embedding-based EA methods have been proposed, such as TransE \citep{bordes2013translating}, GCN \citep{kipf2016semi}, GAT \citep{velivckovic2017graph}, and their variants \citep{sun2017cross,zhu2021relation}. 
All embedding-based EA methods adopt the information from graph structures \citep{chen2017multilingual}, while some methods utilize textual information \citep{xu2019cross,wu2019relation,zhu2021raga} or entity attributes \citep{sun2017cross,trisedya2019entity}.


\noindent \textbf{Global Entity Alignment: }
%
Only matching entities with its nearest neighbor is a ``greedy'' scheme, which is difficult to find the global optimal pair-matching for all the entities. Besides, one mismatched entity pair can disturb the greedy matching of the rest entities, and the match error can be accumulated along the greedy matching process. Therefore, global entity alignment, which considers all alignments jointly,
have recently attracted considerable attention \citep{xu2020coordinated,zeng2020collective,zhu2021raga,lin2021echoea}. However, globally solving optimal matching costs much higher complexity. To improve the efficiency, \citet{xu2020coordinated} decomposes the entire search space into many isolated sub-spaces and consequently restricts the cross-subspace alignment.  \citet{zeng2020collective}  reduces the complexity by requiring the entity pairs to be stable matches and using deferred acceptance algorithm \citep{mullin1952matching,roth2008deferred} to find the alignments.






\vspace{-1mm}
\section{Experimental Setups}
\vspace{-1mm}
\subsection{Datasets and Evaluation}

For the entity alignment task, we follow the prior work~\citep{sun2017cross,wu2019relation,zhu2021raga} and test our method on the DBP15K dataset~\citep{sun2017cross}. For the dangling entity detection task, \citet{sun2021knowing} construct a DBP2.0 dataset with dangling entity labels. However, the DBP2.0 dataset does not contain textual entity name of each entity, which is incompatible for our method to extract pseudo entity pairs. Therefore, we construct a new MedED dataset to support DED task. The details are described below:


\textbf{MedED Dataset Construction: }
The Unified Medical Language System (UMLS) \citep{lindberg1993unified} is a large-scale resource containing over 4 million unique medical concepts and over 87 million relation triples. Concepts in UMLS have several terms in different languages. We extract concepts that contain terms in the selected language as entities to construct new monolingual KG, and retain the relations between entities. For the entity names, we select the preferred terms in UMLS. The criterion of entity pairs is whether entities belong to the same concept. An dangling entity is set  if its original concept is not in the other KG. We extracted the KGs of English, French, and Spanish and then constructed the KG pairs of FR-EN (French to English) and ES-EN (Spanish to English). We select 20,000 entities with the most relation triples in UMLS for the specified language, and then drop the entities unrelated to other selected entities. Supplementary~\ref{sec:data statistics} shows the statistics of the new dataset, MedED. For both EA and DED tasks, we split 70\% of the data as the testing set. Even though our method does not rely on the training set, we keep the remaining 30\% as the training set for further model comparison and ablation study.

\textbf{DBP15K Dataset:}  DBP15K~\citep{sun2017cross} contains three pairs of cross-lingual KGs, ZH-EN (Chinese to English), JA-EN (Japanese to English), and FR-EN (French to English). Each KG contains approximately 20 thousand entities, and every KG pair contains 15 thousand entity pairs (Table~\ref{tab:data statistics}). Following the setting in previous works \citep{sun2017cross,wu2019relation,zhu2021raga}, we keep 70\% of entity pairs for testing and 30\% for training.

\textbf{Evaluation Metrics: } For entity alignment, we  follow previous works \citep{xu2020coordinated,zeng2020collective} and compute Hits@1 score,  which  indicates the percentage of the targets that have been correctly ranked in the top-$1$. The prior works \citep{sun2017cross,sun2018bootstrapping,wu2019relation,zhu2021raga} compute Hits$\widetilde{\text{@}}$1 in a \emph{relaxed setting} in which only the entities in testing pairs are taken into account, assuming that any source entity has a counterpart in the target KG.  To the relaxed evaluation, we also compute Hits$\widehat{\text{@}}$1 in a \emph{practical setting} in which for every testing entity, the list of candidate counterparts consists of all entities in the other KG.
For dangling entity detection, we compute the classification precision, recall, and F1-score as evaluation measurements.
\vspace{-2mm}
\subsection{Baselines}
\vspace{-1mm}
For entity alignment, we compare our method with following baselines: (1) Init-Emb, the entity textual embeddings used in SoTead; (2) the supervised EA methods: MTransE \citep{chen2017multilingual}, JAPE \citep{sun2017cross}, BootEA \citep{sun2018bootstrapping}, RDGCN \citep{wu2019relation}, CEA \citep{zeng2020collective}, RNM \citep{zhu2021relation}, RAGA \citep{zhu2021raga}, SelfKG \citep{liu2021self}, and EchoEA \citep{lin2021echoea}; (3) the unsupervised EA methods: SelfKG \citep{liu2021self}, UEA \citep{zeng2021towards}, and SEU \citep{mao2021alignment}. Here the CEA, RAGA, and EchoEA use the deferred acceptance algorithm (DAA) to globally align entities.

Our proposed method is also compatible with supervised training entity pairs, so we provide both unsupervised and supervised versions of our method: (1) the unsupervised method, SoTead, described in Section~\ref{sec:proposed-method}. (2) the supervised version of SoTead, which combines the training entity pairs and the pseudo entity pairs for the alignment loss, denoted as SoTead*. 

\vspace{-2mm}
\subsection{Implementation Details}
\vspace{-1mm}
Following \citet{wu2019relation}, we translate entity names in MedED to English via Google Translate and then use mean of word vector from GloVe \citep{pennington2014glove} to obtain textual entity embeddings. For entities in DBP15K, we inherit the initial embeddings used in previous works \citep{wu2019relation,zeng2021towards,zhu2021raga,zhu2021relation,lin2021echoea}. The threshold for pseudo pairs $\varepsilon$ is 0.99, and $N=3$ in globally guided refining loss. The initial value of $w(t)$ is 0.3 and $w(t)$ decreases linearly to 0 at 1/4 of the total training steps. We follow GNN setups from RAGA~\citep{zhu2021raga} to encode the enhanced entity embeddings.
The default value of $K$ is set to 100. We grid search 100 paired quantiles of $\{l_{i}^u, 1\leq i \leq m\}$ and $\{l_{j}^v, 1\leq j \leq n\}$ with $K=10$ to obtain $\alpha^*$ and $\beta^*$.

\begin{table}[t]
  \setlength\tabcolsep{1.2pt}
  \small
  \centering
    \begin{tabular}{lccccc}
    \toprule
     &
      \multicolumn{3}{c}{DBP15K} &
      \multicolumn{2}{c}{MedED}
      \\
     &
      \multicolumn{1}{c}{ZH-EN} &
      \multicolumn{1}{c}{JA-EN} &
      \multicolumn{1}{c}{FR-EN} &
      \multicolumn{1}{c}{FR-EN} &
      \multicolumn{1}{c}{SP-EN}
      \\
    \midrule
    \underline{Init-Emb} &
      57.5  &
      65.0  &
      81.8  &
      71.6  &
      68.5
      \\
     \midrule
    \textbf{Supervised } &
       &
       &
       &
       &
      \\
     \midrule
    MTransE &
      30.8  &
      27.9  &
      24.4  &
      -  &  -
      \\
    JAPE &
      73.1  &
      82.8  &
      -  &
      -  &  -
      \\
    BootEA &
      62.9  &
      62.2  &
      65.3  &
      -  &  -
      \\
    \underline{RDGCN} &
      70.8  &
      76.7  &
      88.6  &
      -  &  -
      \\
    \underline{RNM} &
      84.0  &
      87.2  &
      93.8  &
      -  &  -
      \\
    GM-EHD-JEA &
      73.6  &
      79.2  &
      92.4  &
      -  &  -
      \\
    CEA &
      78.7  &
      86.3  &
      97.2  &
      -  &  -
      \\
    \underline{RAGA} &
      87.3  &
      90.9  &
      96.6  &
      96.2  &
      97.0
      \\
    \underline{EchoEA} &
      89.1  &
      93.2  &
      \textbf{98.9} &
      -  &  -
      \\
    \underline{SoTead*} &
      \textbf{91.5} &
      \textbf{94.1} &
      98.4  &
      \textbf{97.4}  &
      \textbf{97.9}
      \\
    \midrule
    \textbf{Unsupervised} &
       &
       &
       &
       &
      \\
    \midrule
    \underline{SelfKG} &
      82.9  &
      89.0  &
      95.9  &
      -  &  -
      \\
    \underline{SoTead} &
      \textbf{87.7}  &
      \textbf{91.5}  &
      \textbf{97.5}  &
      97.0  &
      97.6
       \\
    \hdashline
      UEA(word+string) &
      91.3  &
      94.0  &
      95.3  &
      -  &  -
      \\
      \underline{SEU}(word+char) &
      90.0  &
      95.6  &
      98.8  &
      -  &  -
      \\
      \underline{SoTead}(word+char) &
      \textbf{93.2}  &
      \textbf{96.8}  &
      \textbf{99.5}  &
      -  &  -
      \\
    \bottomrule
    \end{tabular}%
  \caption{The results of Hits$\widetilde{\text{@}}$1 on EA task on  DBP15K and MedED (relaxed setting). The underlined models use the same initial entity embeddings. 
  The CEA, RAGA and EchoEA use the DAA for global alignment. The dashed line split the unsupervised part into model merely using word embedding and using word embedding and char-level or string-level information.}
  \label{tab:Table 2}%
  \label{tab:hit1-EA}
  \vspace{-2mm}
\end{table}%

\begin{table}[t]
  \setlength\tabcolsep{1.9pt}
  \small
  \centering
    \begin{tabular}{lcccccccc}
    \toprule
     &
      \multicolumn{4}{c}{FR-EN} &
      \multicolumn{4}{c}{ES-EN}
      \\
    \midrule
     &
      EA &
      \multicolumn{3}{c}{DED} &
      EA &
      \multicolumn{3}{c}{DED}
      \\
     &
      H$\widehat{\text{@}}$1 &
      P &
      R &
      F &
      H$\widehat{\text{@}}$1 &
      P &
      R &
      F
      \\
    \midrule
    RAGA &
      78.7  &
      - &
      - &
      - &
      82.7  &
      - &
      - &
      -
      \\
    SoTead(DAA) &
      77.4  &
      - &
      - &
      - &
      87.0  &
      - &
      - &
      -
      \\
    Distance &
      - &
      78.1  &
      73.4  &
      75.7  &
      - &
      78.6  &
      \textbf{86.1} &
      82.2 
      \\
    SoTead &
       &
       &
       &
       &
       &
       &
       &
      
      \\
    {\ \ \ \ \ K=1} &
      79.8  &
      96.1  &
      \textbf{79.4} &
      \textbf{86.9} &
      86.0  &
      90.4  &
      84.2  &
      \textbf{87.2}
      \\
    {\ \ \ \ \ K=10} &
      80.3  &
      96.3  &
      75.3  &
      84.5  &
      87.4  &
      93.5  &
      68.4  &
      79.0 
      \\
    {\ \ \ \ \ K=100} &
      80.5  &
      96.4  &
      74.8  &
      84.2  &
      87.7  &
      93.3  &
      64.6  &
      76.4 
      \\
    SoTead* &
      \textbf{82.6} &
      \textbf{97.6} &
      65.4  &
      78.3  &
      \textbf{90.1} &
      \textbf{94.1} &
      69.4  &
      79.9 
      \\
    \bottomrule
    \end{tabular}%
  \caption{EA and DED results on MedED (practical setting). H$\widehat{\text{@}}$1, P, R, and F denotes Hits$\widehat{\text{@}}$1, precision, recall, and F-score. $K = 1,10,100$ refers to the proposed global alignment method that keeps the top $K (= 1, 10, 100)$ rank similarity entities for each entity. The SoTead(DAA) and RAGA use the DAA for global alignment.}
  \label{tab:Table 3}%
  \vspace{-2mm}
\end{table}%

\vspace{-2mm}
\section{Experimental Results}
\vspace{-1mm}
\subsection{Entity Alignment Results}
Table~\ref{tab:hit1-EA} shows the results of EA on DBP15K and MedED datasets. Following the previous work, we adopt the relaxed evaluation setting. The results with practical evaluation settings for Hits$\widehat{\text{@}}$1 are listed in the Supplementary~\ref{sec:appendixA1}. 

When compared under \textbf{supervised} setups,   the SoTead achieves comparable results on DBP15K.  Without supervision, the SoTead outperforms all competing methods except the EchoEA method. Besides, the gap between SoTead and EchoEA is marginal. In addition, SoTead* outperforms all methods in MedED.
When compared under \textbf{unsupervised} setups, SoTead outperforms all other unsupervised baselines.
Besides, UEA and SEU use the extra string-level and char-level information from entity names respectively to enhance the models, denoted as UEA(word+string) and SEU(word+char). For fair comparison, we follow SEU and add the char-level distance to the cost between entities cross KGs. The  the char-level distance is the Manhattan distance between bi-gram vectors from translated entity names used in SEU. This model is denoted as SoTead(word+char) and is superior to UEA and SEU.

\vspace{-1mm}
\subsection{Dangling Entity Detection Results}
Table~\ref{tab:Table 3} shows the results of EA and DED on MedED. Note that global alignment with DED should consider all entities. We select the practical setting in the EA evaluation. 

As shown in Table~\ref{tab:Table 3}, for the EA task, SoTead achieves better results compared to the supervised RAGA and the variants of our method with DAA. 
For the DED task, our method focuses more on the precision in recognizing dangling entities. The results of SoTead and SoTead* are also much better than the Distance, which is the same as the threshold-based unsupervised attempt of \citet{zhao2020experimental} on DED by searching the best threshold on the dangling training set. These results indicate that SoTead successfully uses unsupervised EA to enhance DED, since DED with high precision reduces the scope of EA and enhances the performance.


\vspace{-2mm}
\subsection{Empirical Runtime Analysis}
\label{sec:section 5.3}
\vspace{-1mm}
Choosing an appropriate $K$ can effectively reduce the time complexity of the semi-constraint OT, since the solving process of the semi-constraint OT could be finished in less than 7, 60, and 5,00 seconds for $K=1,10,100$ in MedED. Furthermore, the results with different $K$ in Table~\ref{tab:Table 3} show that the performance of $K=10$ and $K=100$ is similar, implying that much larger value of $K$ may not bring significant improvement and $K=100$ is enough for the proposed method considering the time consuming.


\vspace{-2mm}
\subsection{Ablation Study}
\vspace{-1mm}
In Table~\ref{tab:Table 4}, we test variants by removing the weight decreasing mechanism of the globally guided refining loss $\mathcal{L}_g$ and the $\mathcal{L}_g$ from SoTead. Besides, in Table~\ref{tab:Table 4}, we replace the proposed semi-constraint OT with DAA. Table~\ref{tab:Table 5} provides other variants in practical setting.  Our main observations are:

\begin{table}[t]
  \setlength\tabcolsep{5pt}
  \small
  \centering
    \begin{tabular}{lccccc}
    \toprule
     &
      \multicolumn{3}{c}{DBP15K} &
      \multicolumn{2}{c}{MedED}
      \\
     &
      ZH &
      JA &
      FR &
      FR &
      ES
      \\
    \midrule
    SoTead &
      87.7  &
     91.5  &
      97.5  &
      97.0  &
      97.6 
      \\
    {\ \ \ \ \ w/o OT} &
      77.9  &
      82.0  &
      92.1  &
      89.5  &
      89.3 
      \\
    {\ \ \ \ \ w/o dec.} &
      87.3  &
      91.0  &
      97.3  &
      96.9  &
      97.3 
      \\
    {\ \ \ \ \ w/o $\mathcal{L}_g$} &
      87.5  &
      91.0  &
      97.3  &
      97.1  &
      97.5 
      \\
     
    SoTead(DAA) &
      84.7  &
      89.1  &
      96.2  &
      95.5  &
      95.6 
      \\
    \bottomrule
    \end{tabular}%
  \caption{Hits$\widetilde{\text{@}}$1 results of method variants (relaxed setting) in the EA task. The OT refers to the semi-constraint optimal transport module. The dec. is the weight decreasing mechanism of the globally guided loss, $\mathcal{L}_g$. ZH, JA, FR and ES denotes the KG pairs ZH-EN, JA-EN, FR-EN and ES-EN.}
  \label{tab:Table 4}%
\end{table}%

\begin{table}[t]
  \centering
  \small
    \begin{tabular}{lcccc}
    \toprule
     &
      \multicolumn{2}{c}{      FR-EN} &
      \multicolumn{2}{c}{     ES-EN}
      \\
     &
      EA &
      DED &
      EA &
      DED
      \\
    \midrule
    SoTead &
      80.3 &
      84.5 &
      87.4 &
      79.0
      \\
    {\ \ \ \ \ w/o virtual} &
      55.5 &
      - &
      65.2 &
      -
      \\
    {\ \ \ \ \ w. golden $\alpha, \beta$} &
      80.9 &
      80.3 &
      87.4 &
      79.0
      \\
    SoTead(CODER) &
      88.4  &
      86.3  &
      93.3  &
      86.5 
      \\
    \bottomrule
    \end{tabular}%
    \vspace{-1mm}
  \caption{Results of method variants (practical setting) in MedED. We report Hits$\widehat{\text{@}}$1 and F-score for EA and DED. The w/o virtual denotes the semi-constraint OT without the virtual entities. The w. golden $\alpha, \beta$ denote that the $\alpha$ and $\beta$ in the semi-constraint OT are selected by the EA training set. SoTead(CODER) replaces  Glove embeddings with a medical language model.}
  \label{tab:Table 5}%
  \vspace{-2mm}
\end{table}%

\begin{table}[t]
  \centering
  \small
    \begin{tabular}{lccc}
    \toprule
     &
      \multicolumn{1}{c}{ZH-EN} &
      \multicolumn{1}{c}{JA-EN} &
      \multicolumn{1}{c}{FR-EN}
      \\
     \hline
    Glove(initial) &
      57.0  &
      63.3  &
      80.7 
      \\
    Glove(SoTead)&
      84.7  &
      89.0  &
      96.6 
      \\
      \hline
    word2vec(initial) &
      44.6  &
      49.6  &
      59.4 
      \\
    word2vec(SoTead) &
      65.0  &
      70.6  &
      79.6 
      \\
      \hline
    FastText(initial) &
      49.7  &
      54.3  &
      64.3 
      \\
    FastText(SoTead)&
      74.7  &
      77.3  &
      88.1 
      \\
      \hline
    MUSE(initial) &
      57.9  &
      61.1  &
      76.6 
      \\
    MUSE(SoTead) &
      83.3  &
      87.0  &
      95.8 
      \\
     \bottomrule
    \end{tabular}%
    \caption{The EA results of Hits$\widehat{\text{@}}$1 on DBP15K  for SoTead trained with different resources of embeddings.  ``initial'' denotes the entity embeddings initialized by taking the mean of word embeddings. }
  \label{tab:Table embeddings}%
\end{table}%

\begin{figure}[t]
    \centering
    \includegraphics[width=\columnwidth]{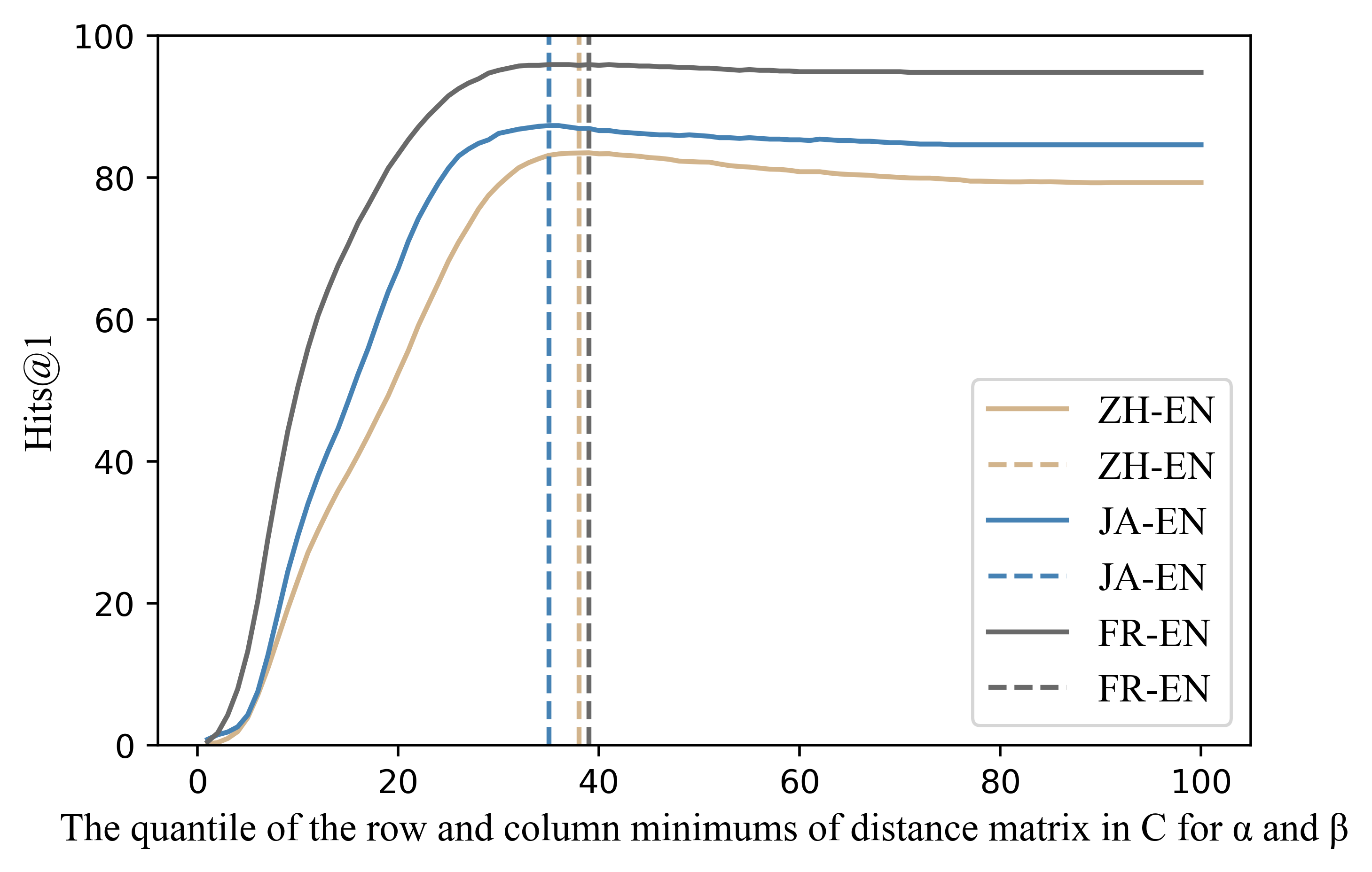}
    \vspace{-3mm}
    \caption{The Hits$\widehat{\text{@}}$1 in DBP15K (practical setting) for the SoTead according to different value selection of the $\alpha$ and $\beta$. The solid line denotes the result of SoTead and dashed line in corresponding color denotes $\alpha^*$ and $\beta^*$ selected by the proposed  selection strategy. }
    \label{fig:Figure grid search}
    \vspace{-1mm}
\end{figure}

\begin{figure}[t]
    \centering
    \includegraphics[width=0.93\columnwidth]{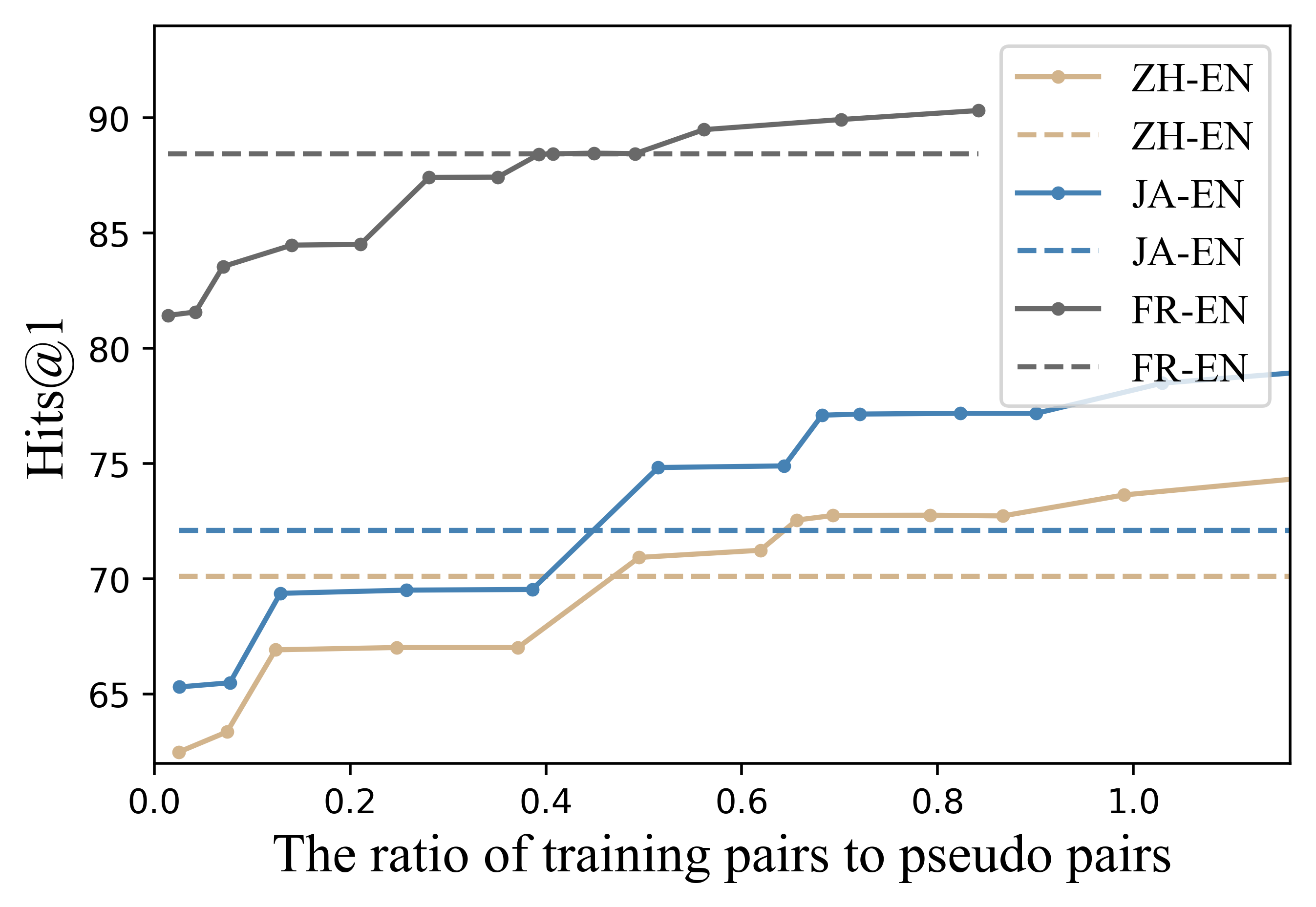}
    \vspace{-3mm}
    \caption{The Hits$\widehat{\text{@}}$1 in DBP15K (practical setting) for the SoTead without $\mathcal{L}_g$ and semi-constraint OT. The solid line and dashed line denotes the method trained with the training entity pairs and pseudo entity pairs, respectively. }
    \label{fig:Figure 2}
    \vspace{-1mm}
\end{figure}

1. The semi-constraint OT is stable and effective, causing significant improvements (4.6$\sim$9.8 Hits$\widehat{\text{@}}$1) compared with the SoTead without semi-constraint OT (Table~\ref{tab:Table 4}). 

2. Ablating the globally guided refining similarity $\mathcal{L}_g$ and the weight decreasing mechanism of $\mathcal{L}_g$ leads to a decrease in most cases (Table~\ref{tab:Table 4}), indicating they are usually helpful.

3. Introducing the virtual entity is necessary. In Table~\ref{tab:Table 5}, the SoTead without virtual entity gains a remarkable decrease on EA from 80.3 to 55.5 in FR-EN and from 87.4 to 65.2 in ES-EN, and cannot be applied to the DAD task.

4. Figure~\ref{fig:Figure grid search} shows that the proposed method for searching proper $\alpha^*$ and $\beta^*$ almost reaches the best Hits$\widehat{\text{@}}$1. Similar conclusion is shown in Table~\ref{tab:Table 5} with the marginal difference between the result of $\alpha^*$ and $\beta^*$ and the golden selection of $\alpha$ and $\beta$ based on the EA training entity pairs.

5. The proportion of how many pseudo entity pairs can play an equal role as true entity pairs changes depending on the KG pairs in Figure~\ref{fig:Figure 2}, but it is still valid to obtain pseudo entity pairs when true entity pairs are unavailable since the proportion is not very low.


To further understand the influence of the quality of initial embeddings, we replace the GloVe in MedED with a pretrained medical language model (LM), the English version of CODER \citep{yuan2020coder}, and show that a proper domain-specific LM trained on a large KG can bring better results (in Table~\ref{tab:Table 5}).
Moreover, we test the effect of different initial word embedding on SoTead performance. We use Glove, word2vec, FastText, and MUSE (in Table~\ref{tab:Table embeddings}). The MUSE is a multilingual word embeddings method, and we generate the initial entity embeddings directly without machine translation. Table~\ref{tab:Table embeddings} indicates that the performance of SoTead relies on the quality of the textual information, since the Hits$\widehat{\text{@}}$1 of SoTead highly related to the Hits$\widehat{\text{@}}$1 of the initial embeddings.

\vspace{-1.5mm}
\section{Conclusion}
\vspace{-1.5mm}
We propose a novel unsupervised method, SoTead, for  entity alignment and dangling entity detection via a semi-constraint optimal transport. By introducing virtual entities on KGs, our method relaxes the alignment constraints, and jointly detects the dangling entities and matchable entity pairs. In experiments, our method reaches state-of-the-art performance and supports the point that  dangling entity detection enhances the entity alignment models. Besides, we build a new cross-lingual medical knowledge graph dataset, MedED, which supports both entity alignment and dangling entity detection tasks. We hope this work provide insight for real-world knowledge graph processing.



\bibliography{anthology,custom}

\begin{thebibliography}{49}
\expandafter\ifx\csname natexlab\endcsname\relax\def\natexlab#1{#1}\fi

\bibitem[{Alvarez-Melis and Jaakkola(2018)}]{alvarez2018gromov}
David Alvarez-Melis and Tommi~S Jaakkola. 2018.
\newblock Gromov-wasserstein alignment of word embedding spaces.
\newblock \emph{arXiv preprint arXiv:1809.00013}.

\bibitem[{Balas et~al.(1993)Balas, Ceria, and Cornu{\'e}jols}]{balas1993lift}
Egon Balas, Sebasti{\'a}n Ceria, and G{\'e}rard Cornu{\'e}jols. 1993.
\newblock A lift-and-project cutting plane algorithm for mixed 0--1 programs.
\newblock \emph{Mathematical programming}, 58(1):295--324.

\bibitem[{Bordes et~al.(2013)Bordes, Usunier, Garcia-Duran, Weston, and
  Yakhnenko}]{bordes2013translating}
Antoine Bordes, Nicolas Usunier, Alberto Garcia-Duran, Jason Weston, and Oksana
  Yakhnenko. 2013.
\newblock Translating embeddings for modeling multi-relational data.
\newblock \emph{Advances in neural information processing systems}, 26.

\bibitem[{Cao et~al.(2019)Cao, Wang, He, Hu, and Chua}]{cao2019unifying}
Yixin Cao, Xiang Wang, Xiangnan He, Zikun Hu, and Tat-Seng Chua. 2019.
\newblock Unifying knowledge graph learning and recommendation: Towards a
  better understanding of user preferences.
\newblock In \emph{The world wide web conference}, pages 151--161.

\bibitem[{Chen et~al.(2018)Chen, Dai, Tao, Shen, Gan, Zhang, Zhang, and
  Carin}]{chen2018adversarial}
Liqun Chen, Shuyang Dai, Chenyang Tao, Dinghan Shen, Zhe Gan, Haichao Zhang,
  Yizhe Zhang, and Lawrence Carin. 2018.
\newblock Adversarial text generation via feature-mover's distance.
\newblock \emph{arXiv preprint arXiv:1809.06297}.

\bibitem[{Chen et~al.(2019)Chen, Zhang, Zhang, Tao, Gan, Zhang, Li, Shen, Chen,
  and Carin}]{chen2019improving}
Liqun Chen, Yizhe Zhang, Ruiyi Zhang, Chenyang Tao, Zhe Gan, Haichao Zhang, Bai
  Li, Dinghan Shen, Changyou Chen, and Lawrence Carin. 2019.
\newblock Improving sequence-to-sequence learning via optimal transport.
\newblock \emph{arXiv preprint arXiv:1901.06283}.

\bibitem[{Chen et~al.(2017)Chen, Tian, Yang, and
  Zaniolo}]{chen2017multilingual}
Muhao Chen, Yingtao Tian, Mohan Yang, and Carlo Zaniolo. 2017.
\newblock Multilingual knowledge graph embeddings for cross-lingual knowledge
  alignment.
\newblock In \emph{Proceedings of the 26th International Joint Conference on
  Artificial Intelligence}, pages 1511--1517.

\bibitem[{Dantzig(2016)}]{dantzig2016linear}
George Dantzig. 2016.
\newblock \emph{Linear programming and extensions}.
\newblock Princeton university press.

\bibitem[{Dantzig et~al.(1955)Dantzig, Orden, Wolfe
  et~al.}]{dantzig1955generalized}
George~B Dantzig, Alex Orden, Philip Wolfe, et~al. 1955.
\newblock The generalized simplex method for minimizing a linear form under
  linear inequality restraints.
\newblock \emph{Pacific Journal of Mathematics}, 5(2):183--195.

\bibitem[{Gal{\'a}rraga et~al.(2017)Gal{\'a}rraga, Razniewski, Amarilli, and
  Suchanek}]{galarraga2017predicting}
Luis Gal{\'a}rraga, Simon Razniewski, Antoine Amarilli, and Fabian~M Suchanek.
  2017.
\newblock Predicting completeness in knowledge bases.
\newblock In \emph{Proceedings of the tenth acm international conference on web
  search and data mining}, pages 375--383.

\bibitem[{Gass(2003)}]{gass2003linear}
Saul~I Gass. 2003.
\newblock \emph{Linear programming: methods and applications}.
\newblock Courier Corporation.

\bibitem[{Jin et~al.(2021)Jin, Yuan, Xiong, Yu, Tan, Chen, Huang, Liu, and
  Yu}]{jin2021biomedical}
Qiao Jin, Zheng Yuan, Guangzhi Xiong, Qianlan Yu, Chuanqi Tan, Mosha Chen,
  Songfang Huang, Xiaozhong Liu, and Sheng Yu. 2021.
\newblock Biomedical question answering: A comprehensive review.
\newblock \emph{arXiv preprint arXiv:2102.05281}.

\bibitem[{Kipf and Welling(2016)}]{kipf2016semi}
Thomas~N Kipf and Max Welling. 2016.
\newblock Semi-supervised classification with graph convolutional networks.
\newblock \emph{arXiv preprint arXiv:1609.02907}.

\bibitem[{Kusner et~al.(2015)Kusner, Sun, Kolkin, and
  Weinberger}]{kusner2015word}
Matt Kusner, Yu~Sun, Nicholas Kolkin, and Kilian Weinberger. 2015.
\newblock From word embeddings to document distances.
\newblock In \emph{International conference on machine learning}, pages
  957--966. PMLR.

\bibitem[{Lin et~al.(2021)Lin, Song, Luo et~al.}]{lin2021echoea}
Xueyuan Lin, Wenyu Song, Haoran Luo, et~al. 2021.
\newblock Echoea: Echo information between entities and relations for entity
  alignment.
\newblock \emph{arXiv preprint arXiv:2107.03054}.

\bibitem[{Lindberg et~al.(1993)Lindberg, Humphreys, and
  McCray}]{lindberg1993unified}
Donald~AB Lindberg, Betsy~L Humphreys, and Alexa~T McCray. 1993.
\newblock The unified medical language system.
\newblock \emph{Yearbook of Medical Informatics}, 2(01):41--51.

\bibitem[{Liu et~al.(2021)Liu, Hong, Wang, Chen, Kharlamov, Dong, and
  Tang}]{liu2021self}
Xiao Liu, Haoyun Hong, Xinghao Wang, Zeyi Chen, Evgeny Kharlamov, Yuxiao Dong,
  and Jie Tang. 2021.
\newblock A self-supervised method for entity alignment.
\newblock \emph{arXiv preprint arXiv:2106.09395}.

\bibitem[{Mao et~al.(2020)Mao, Wang, Xu, Wu, and Lan}]{mao2020relational}
Xin Mao, Wenting Wang, Huimin Xu, Yuanbin Wu, and Man Lan. 2020.
\newblock Relational reflection entity alignment.
\newblock In \emph{Proceedings of the 29th ACM International Conference on
  Information \& Knowledge Management}, pages 1095--1104.

\bibitem[{Mao et~al.(2021)Mao, Wang, Wu, and Lan}]{mao2021alignment}
Xinnian Mao, Wenting Wang, Yuanbin Wu, and Man Lan. 2021.
\newblock From alignment to assignment: Frustratingly simple unsupervised
  entity alignment.
\newblock In \emph{Proceedings of the 2021 Conference on Empirical Methods in
  Natural Language Processing}, pages 2843--2853.

\bibitem[{Mitchell(2002)}]{mitchell2002branch}
John~E Mitchell. 2002.
\newblock Branch-and-cut algorithms for combinatorial optimization problems.
\newblock \emph{Handbook of applied optimization}, 1:65--77.

\bibitem[{Mullin and Stalnaker(1952)}]{mullin1952matching}
FJ~Mullin and John~M Stalnaker. 1952.
\newblock The matching plan for internship placement.
\newblock \emph{J. MED. EDUC.}, 27:193--193.

\bibitem[{Padberg and Rinaldi(1991)}]{padberg1991branch}
Manfred Padberg and Giovanni Rinaldi. 1991.
\newblock A branch-and-cut algorithm for the resolution of large-scale
  symmetric traveling salesman problems.
\newblock \emph{SIAM review}, 33(1):60--100.

\bibitem[{Paulheim(2018)}]{paulheim2018much}
Heiko Paulheim. 2018.
\newblock How much is a triple? estimating the cost of knowledge graph
  creation.

\bibitem[{Pennington et~al.(2014)Pennington, Socher, and
  Manning}]{pennington2014glove}
Jeffrey Pennington, Richard Socher, and Christopher~D Manning. 2014.
\newblock Glove: Global vectors for word representation.
\newblock In \emph{Proceedings of the 2014 conference on empirical methods in
  natural language processing (EMNLP)}, pages 1532--1543.

\bibitem[{Roth(2008)}]{roth2008deferred}
Alvin~E Roth. 2008.
\newblock Deferred acceptance algorithms: History, theory, practice, and open
  questions.
\newblock \emph{international Journal of game Theory}, 36(3):537--569.

\bibitem[{Salimans et~al.(2018)Salimans, Zhang, Radford, and
  Metaxas}]{salimans2018improving}
Tim Salimans, Han Zhang, Alec Radford, and Dimitris Metaxas. 2018.
\newblock Improving gans using optimal transport.
\newblock \emph{arXiv preprint arXiv:1803.05573}.

\bibitem[{Savenkov and Agichtein(2016)}]{savenkov2016crqa}
Denis Savenkov and Eugene Agichtein. 2016.
\newblock Crqa: Crowd-powered real-time automatic question answering system.
\newblock In \emph{Fourth AAAI conference on human computation and
  crowdsourcing}.

\bibitem[{Sun et~al.(2021)Sun, Chen, and Hu}]{sun2021knowing}
Zequn Sun, Muhao Chen, and Wei Hu. 2021.
\newblock Knowing the no-match: Entity alignment with dangling cases.

\bibitem[{Sun et~al.(2017)Sun, Hu, and Li}]{sun2017cross}
Zequn Sun, Wei Hu, and Chengkai Li. 2017.
\newblock Cross-lingual entity alignment via joint attribute-preserving
  embedding.
\newblock In \emph{International Semantic Web Conference}, pages 628--644.
  Springer.

\bibitem[{Sun et~al.(2018)Sun, Hu, Zhang, and Qu}]{sun2018bootstrapping}
Zequn Sun, Wei Hu, Qingheng Zhang, and Yuzhong Qu. 2018.
\newblock Bootstrapping entity alignment with knowledge graph embedding.
\newblock In \emph{Proceedings of the 27th International Joint Conference on
  Artificial Intelligence}, pages 4396--4402.

\bibitem[{Trisedya et~al.(2019)Trisedya, Qi, and Zhang}]{trisedya2019entity}
Bayu~Distiawan Trisedya, Jianzhong Qi, and Rui Zhang. 2019.
\newblock Entity alignment between knowledge graphs using attribute embeddings.
\newblock In \emph{Proceedings of the AAAI Conference on Artificial
  Intelligence}, volume~33, pages 297--304.

\bibitem[{Veli{\v{c}}kovi{\'c} et~al.(2017)Veli{\v{c}}kovi{\'c}, Cucurull,
  Casanova, Romero, Lio, and Bengio}]{velivckovic2017graph}
Petar Veli{\v{c}}kovi{\'c}, Guillem Cucurull, Arantxa Casanova, Adriana Romero,
  Pietro Lio, and Yoshua Bengio. 2017.
\newblock Graph attention networks.
\newblock \emph{arXiv preprint arXiv:1710.10903}.

\bibitem[{Villani(2009)}]{villani2009optimal}
C{\'e}dric Villani. 2009.
\newblock \emph{Optimal transport: old and new}, volume 338.
\newblock Springer.

\bibitem[{Wang et~al.(2018)Wang, Lv, Lan, and Zhang}]{wang2018cross}
Zhichun Wang, Qingsong Lv, Xiaohan Lan, and Yu~Zhang. 2018.
\newblock Cross-lingual knowledge graph alignment via graph convolutional
  networks.
\newblock In \emph{Proceedings of the 2018 Conference on Empirical Methods in
  Natural Language Processing}, pages 349--357.

\bibitem[{Wolsey and Nemhauser(1999)}]{wolsey1999integer}
Laurence~A Wolsey and George~L Nemhauser. 1999.
\newblock \emph{Integer and combinatorial optimization}, volume~55.
\newblock John Wiley \& Sons.

\bibitem[{Wu et~al.(2019)Wu, Liu, Feng, Wang, Yan, and Zhao}]{wu2019relation}
Yuting Wu, Xiao Liu, Yansong Feng, Zheng Wang, Rui Yan, and Dongyan Zhao. 2019.
\newblock Relation-aware entity alignment for heterogeneous knowledge graphs.
\newblock \emph{arXiv preprint arXiv:1908.08210}.

\bibitem[{Xiong et~al.(2017)Xiong, Power, and Callan}]{xiong2017explicit}
Chenyan Xiong, Russell Power, and Jamie Callan. 2017.
\newblock Explicit semantic ranking for academic search via knowledge graph
  embedding.
\newblock In \emph{Proceedings of the 26th international conference on world
  wide web}, pages 1271--1279.

\bibitem[{Xu et~al.(2020)Xu, Song, Feng, Song, and Yu}]{xu2020coordinated}
Kun Xu, Linfeng Song, Yansong Feng, Yan Song, and Dong Yu. 2020.
\newblock Coordinated reasoning for cross-lingual knowledge graph alignment.
\newblock In \emph{Proceedings of the AAAI Conference on Artificial
  Intelligence}, volume~34, pages 9354--9361.

\bibitem[{Xu et~al.(2019)Xu, Wang, Yu, Feng, Song, Wang, and Yu}]{xu2019cross}
Kun Xu, Liwei Wang, Mo~Yu, Yansong Feng, Yan Song, Zhiguo Wang, and Dong Yu.
  2019.
\newblock Cross-lingual knowledge graph alignment via graph matching neural
  network.
\newblock In \emph{Proceedings of the 57th Annual Meeting of the Association
  for Computational Linguistics}, pages 3156--3161.

\bibitem[{Yu et~al.(2017)Yu, Yin, Hasan, dos Santos, Xiang, and
  Zhou}]{yu2017improved}
Mo~Yu, Wenpeng Yin, Kazi~Saidul Hasan, Cicero dos Santos, Bing Xiang, and Bowen
  Zhou. 2017.
\newblock Improved neural relation detection for knowledge base question
  answering.
\newblock In \emph{Proceedings of the 55th Annual Meeting of the Association
  for Computational Linguistics (Volume 1: Long Papers)}, pages 571--581.

\bibitem[{Yuan et~al.(2020)Yuan, Zhao, Sun, Li, Wang, and Yu}]{yuan2020coder}
Zheng Yuan, Zhengyun Zhao, Haixia Sun, Jiao Li, Fei Wang, and Sheng Yu. 2020.
\newblock Coder: Knowledge infused cross-lingual medical term embedding for
  term normalization.
\newblock \emph{arXiv preprint arXiv:2011.02947}.

\bibitem[{Zeng et~al.(2021{\natexlab{a}})Zeng, Li, Hou, Li, and
  Feng}]{zeng2021comprehensive}
Kaisheng Zeng, Chengjiang Li, Lei Hou, Juanzi Li, and Ling Feng.
  2021{\natexlab{a}}.
\newblock A comprehensive survey of entity alignment for knowledge graphs.
\newblock \emph{AI Open}, 2:1--13.

\bibitem[{Zeng et~al.(2021{\natexlab{b}})Zeng, Zhao, Tang, Li, Luo, and
  Zheng}]{zeng2021towards}
Weixin Zeng, Xiang Zhao, Jiuyang Tang, Xinyi Li, Minnan Luo, and Qinghua Zheng.
  2021{\natexlab{b}}.
\newblock Towards entity alignment in the open world: An unsupervised approach.
\newblock \emph{arXiv preprint arXiv:2101.10535}.

\bibitem[{Zeng et~al.(2020)Zeng, Zhao, Tang, and Lin}]{zeng2020collective}
Weixin Zeng, Xiang Zhao, Jiuyang Tang, and Xuemin Lin. 2020.
\newblock Collective entity alignment via adaptive features.
\newblock In \emph{2020 IEEE 36th International Conference on Data Engineering
  (ICDE)}, pages 1870--1873. IEEE.

\bibitem[{Zhang et~al.(2020)Zhang, Liu, Chen, Chen, Liu, Xiang, and
  Zheng}]{zhang2020industry}
Ziheng Zhang, Hualuo Liu, Jiaoyan Chen, Xi~Chen, Bo~Liu, YueJia Xiang, and
  Yefeng Zheng. 2020.
\newblock An industry evaluation of embedding-based entity alignment.
\newblock In \emph{Proceedings of the 28th International Conference on
  Computational Linguistics: Industry Track}, pages 179--189.

\bibitem[{Zhao(2020)}]{zhao2020towards}
Xiang Zhao. 2020.
\newblock Towards knowledge graphs federations: Issues and technologies.
\newblock In \emph{Asia-Pacific Web (APWeb) and Web-Age Information Management
  (WAIM) Joint International Conference on Web and Big Data}, pages 66--79.
  Springer.

\bibitem[{Zhao et~al.(2020)Zhao, Zeng, Tang, Wang, and
  Suchanek}]{zhao2020experimental}
Xiang Zhao, Weixin Zeng, Jiuyang Tang, Wei Wang, and Fabian Suchanek. 2020.
\newblock An experimental study of state-of-the-art entity alignment
  approaches.
\newblock \emph{IEEE Transactions on Knowledge \& Data Engineering}, (01):1--1.

\bibitem[{Zhu et~al.(2021{\natexlab{a}})Zhu, Ma, and Wang}]{zhu2021raga}
Renbo Zhu, Meng Ma, and Ping Wang. 2021{\natexlab{a}}.
\newblock Raga: Relation-aware graph attention networks for global entity
  alignment.
\newblock In \emph{PAKDD (1)}, pages 501--513. Springer.

\bibitem[{Zhu et~al.(2021{\natexlab{b}})Zhu, Liu, Wu, and Du}]{zhu2021relation}
Yao Zhu, Hongzhi Liu, Zhonghai Wu, and Yingpeng Du. 2021{\natexlab{b}}.
\newblock Relation-aware neighborhood matching model for entity alignment.
\newblock In \emph{Proceedings of the AAAI Conference on Artificial
  Intelligence}, volume~35, pages 4749--4756.

\end{thebibliography}
\bibliographystyle{acl_natbib}

\clearpage
\appendix

\section{Supplementary}
\subsection{Branch-and-Cut Algorithm}
\label{sec:branch-and-cut}
We describe the branch-and-cut algorithm \citep{mitchell2002branch} to solve the proposed semi-constraint optimal transport (SOP) as follows:
\begin{algorithm}
	\renewcommand{\algorithmicrequire}{\textbf{Input:}}
	\renewcommand{\algorithmicensure}{\textbf{Output:}}
	\caption{Branch-and-Cut Algorithm}
	\label{alg1}
	\begin{algorithmic}[1]
		\STATE \textbf{Step 1}. Initialization: Denote the initial SOP as $SOP^0$ and the active nodes $L=\{SOP^0\}$. Set the upper bound of the object function $z=\sum_{i=1}^m \sum_{j=1}^n c_{ij}\psi_{ij} + \sum_{i=1}^m c_{i0} \psi_{i0} + \sum_{j=1}^n c_{0j} \psi_{0j}$ to be $\overline{z}=+ \infty$ and set the lower bound of problem $l\in L$ to be $\underline{z}_l = -\infty$.
		\STATE \textbf{Step 2}. Termination: If $L = \varnothing$, then the solution $\psi^{*}$ which yielded the incumbent objective value $\overline{z}$ is optimal. If no such $\psi^{*}$ exists, the SOP is infeasible.
		\STATE \textbf{Step 3} Problem selection: Select and delete a problem $SOP^l$ from $L$.
		\STATE \textbf{Step 4}. Relaxation: Solve the linear programming relaxation of $SOP^l$. If the relaxation is infeasible, set $\underline{z}^l = +\infty$ and go to Step 6. Let $\underline{z}^l$ denote the optimal objective value of the relaxation if it is finite and let $\psi^{lR}$ be an optimal solution; otherwise set $\underline{z}^l = -\infty$.
        \STATE \textbf{Step 5}. Add cutting planes: If desired, search for cutting planes that are violated by $\psi^{lR}$; if any are found, add them to the relaxation and return to Step 4.
        \STATE \textbf{Step 6}. Fathoming and Pruning:
        \STATE\quad\quad (a) If $z^l \geq \overline{z}$, go to Step 2.
        \STATE\quad\quad (b) If $z^l < \overline{z}$ and $\psi^{lR}$ is integral feasible, update $\overline{z}=\underline{z}_l$ and delete all $l \in L$ with $\underline{z}_l \geq \overline{z}$. Go to Step 2.
        \STATE \textbf{Step 7} Partitioning: Let $\{S^{lj}\}_{j=1}^{j=k}$ be a partition of the constraint set $S^l$ of $SOP^l$. Add $\{SOP^{lj}\}_{j=1}^{j=k}$ to $L$, where $SOP^{lj}$ is $SOP^{l}$ with feasible region restricted to $S^{lj}$ and $\underline{z}_{lj}$ for $j=1,...,k$ is set to the value of $z^l$ for the parent problem $l$. Go to Step 2.
	\end{algorithmic}  
\end{algorithm}

\subsection{Data Statistics}
\label{sec:data statistics}
We provide the statistics of MedED and DBP15K in Table~\ref{tab:data statistics}, including entities, types of relations, relations triples, entity pairs, and dangling entities in each KG.
\begin{table}[t]
\setlength\tabcolsep{2pt}
\small
  \centering
    \begin{tabular}{ccccccc}
    \hline
    \multicolumn{2}{c}{Datasets} &
      \multicolumn{1}{c}{\#Ent.} &
      \multicolumn{1}{c}{\#Rel.} &
      \multicolumn{1}{c}{\#Trip.} &
      \multicolumn{1}{c}{\#Pairs} &
      \#Dang.
      \bigstrut\\
    \hline
    \multicolumn{1}{c}{\multirow{2}[1]{*}{MedED}} &
      FR &
      19,382 &
      431 &
      455,368 &
      \multirow{2}[1]{*}{6,365} &
      \multicolumn{1}{c}{13,017}
      \bigstrut[t]\\
     &
      EN &
      18,632 &
      622 &
      841,792 &
       &
      \multicolumn{1}{c}{12,267}
      \\
    \multicolumn{1}{c}{\multirow{2}[0]{*}{MedED}} &
      ES &
      19,228 &
      546 &
      594,130 &
      \multirow{2}[0]{*}{11,153} &
      \multicolumn{1}{c}{8,075}
      \\
     &
      EN &
      18,632 &
      622 &
      841,792 &
       &
      \multicolumn{1}{c}{7,479}
      \\
    \multicolumn{1}{c}{\multirow{2}[0]{*}{DBP15K}} &
      ZH &
      19,388 &
      1,700 &
      70,414 &
      \multirow{2}[0]{*}{15,000} &
      -
      \\
     &
      EN &
      19,572 &
      1,322 &
      95,142 &
       &
      -
      \\
    \multicolumn{1}{c}{\multirow{2}[0]{*}{DBP15K}} &
      JA &
      19,814 &
      1,298 &
      77,214 &
      \multirow{2}[0]{*}{15,000} &
      -
      \\
     &
      EN &
      19,780 &
      1,152 &
      93,484 &
       &
      -
      \\
    \multicolumn{1}{c}{\multirow{2}[1]{*}{DBP15K}} &
      FR &
      19,661 &
      902 &
      105,998 &
      \multirow{2}[1]{*}{15,000} &
      -
      \\
     &
      EN &
      19,993 &
      1,207 &
      115,722 &
       &
      -
      \bigstrut[b]\\
    \hline
    \end{tabular}%
    \caption{Statistics of MedED and DBP15K.}
  \label{tab:data statistics}%
\end{table}%

\subsection{Practiacl Evaluation Results}
\label{sec:appendixA1}
For completeness, this supplementary reports the EA results on DBP15K in practiacl evaluation setting (Table~\ref{tab:Table 6}).We compared our methods with the RAGA, since we adopt the part of graph embedding in RAGA in our framework. 
\begin{table}[ht]
  \setlength\tabcolsep{0.0pt}
  \small
  \centering
    \begin{tabular}{lcrrcrrcrr}
    \toprule
     &
      \multicolumn{3}{c}{ZH-EN} &
      \multicolumn{3}{c}{JA-EN} &
      \multicolumn{3}{c}{FR-EN}
      \\
     &
      \multicolumn{1}{l}{$\widehat{@}1$} &
      \multicolumn{1}{l}{$\widehat{@}10$} &
      \multicolumn{1}{l}{MRR} &
      \multicolumn{1}{l}{$\widehat{@}1$} &
      \multicolumn{1}{l}{$\widehat{@}10$} &
      \multicolumn{1}{l}{MRR} &
      \multicolumn{1}{l}{$\widehat{@}1$} &
      \multicolumn{1}{l}{$\widehat{@}10$} &
      \multicolumn{1}{l}{MRR}
      \\
    \midrule
    Init-Emb &
      57.0  &
      \multicolumn{1}{c}{68.6 } &
      \multicolumn{1}{c}{61.1 } &
      63.3  &
      \multicolumn{1}{c}{75.3 } &
      \multicolumn{1}{c}{67.6 } &
      80.7  &
      \multicolumn{1}{c}{89.0 } &
      \multicolumn{1}{c}{83.5 }
      \\
    RAGA(w/o DAA)&
      72.5  &
      \multicolumn{1}{c}{90.3 } &
      \multicolumn{1}{c}{79.0 } &
      77.3  &
      \multicolumn{1}{c}{93.1 } &
      \multicolumn{1}{c}{82.9 } &
      88.4  &
      \multicolumn{1}{c}{97.2 } &
      \multicolumn{1}{c}{91.7 }
      \\
    SoTead(w/o OT)&
      75.1  &
      \multicolumn{1}{c}{89.2 } &
      \multicolumn{1}{c}{80.2 } &
      79.3  &
      \multicolumn{1}{c}{91.8 } &
      \multicolumn{1}{c}{83.9 } &
      91.1  &
      \multicolumn{1}{c}{97.4 } &
      \multicolumn{1}{c}{93.4 }
      \\
    \midrule
    RAGA &
      83.4  &
       &
       &
      74.2  &
       &
       &
      92.9  &
       &
      
      \\
    SoTead(DAA) &
      79.9  &
       &
       &
      76.9  &
       &
       &
      93.5  &
       &
      
      \\
    SoTead &
      \textbf{84.7}  &
       &
       &
      \textbf{89.0}  &
       &
       &
      \textbf{96.6}  &
       &
      
      \\
    \bottomrule
    \end{tabular}%
  \caption{EA results on DBP15K (practical setting). $\widehat{@}1$ and $\widehat{@}10$ denotes the Hits$\widehat{\text{@}}$1 and Hits$\widehat{\text{@}}$10. In the part of local alignment, RAGA(w/o DAA) and SoTead(w/o OT) refer to the RAGA without DAA and SoTead without OT. The SoTead(DAA) refers to the variant of SoTead by replacing the semi-constraint OT with DAA. MRR is the average of the reciprocal of the rank results.}
  \label{tab:Table 6}%
\end{table}%

\end{document}